\documentclass{article}
\usepackage{ijcai19}

\usepackage{times}
\usepackage{soul}
\usepackage{url}
\usepackage[hidelinks]{hyperref}
\usepackage[utf8]{inputenc}
\usepackage[small]{caption}
\usepackage{graphicx}
\usepackage{amsmath}
\usepackage{booktabs}
\usepackage{amssymb}
\usepackage{multirow}
\usepackage{mathtools}
\usepackage{float}
\usepackage{algorithm}
\usepackage{algorithmic}
\usepackage{setspace}
\title{Video Extrapolation with an Invertible Linear Embedding}

\author{
Robert Pottorff\footnote{Contact Author}\and
Jared T Nielsen\and
David Wingate\\
\affiliations
Brigham Young University\\
\emails
rpottorff@byu.net,
jaredtnielsen@gmail.com,
wingated@cs.byu.edu
}

\begin{document}

\maketitle

\begin{abstract}
We predict future video frames from complex dynamic scenes, using an invertible neural network as the encoder of a nonlinear dynamic system with latent linear state evolution. Our invertible linear embedding (ILE) demonstrates successful learning, prediction and latent state inference. In contrast to other approaches, ILE does not use any explicit reconstruction loss or simplistic pixel-space assumptions. Instead, it leverages invertibility to optimize the likelihood of image sequences exactly, albeit indirectly. Comparison with a state-of-the-art method demonstrates the viability of our approach. 
\end{abstract}


\section{Introduction}

Video frame extrapolation is the generation of future frames conditioned on past ones. Due to the ubiquity of image and video sequences, video prediction plays a central role in diverse fields such as self-driving vehicles and reinforcement learning. In these domains, high quality video prediction correlates directly with improved practical performance.

Frame prediction also offers a well-posed unsupervised objective for representation learning. Any successful algorithm must have extracted salient features useful for describing both the content and dynamics of a scene. To some degree, video prediction and representation learning are essentially the same task. With the right representation, prediction is easy, stable, and efficient; with the wrong one, it may be difficult or impossible \cite{bengio2013representation}.

 \begin{figure}[ht]
\centering
\includegraphics[width=\linewidth]{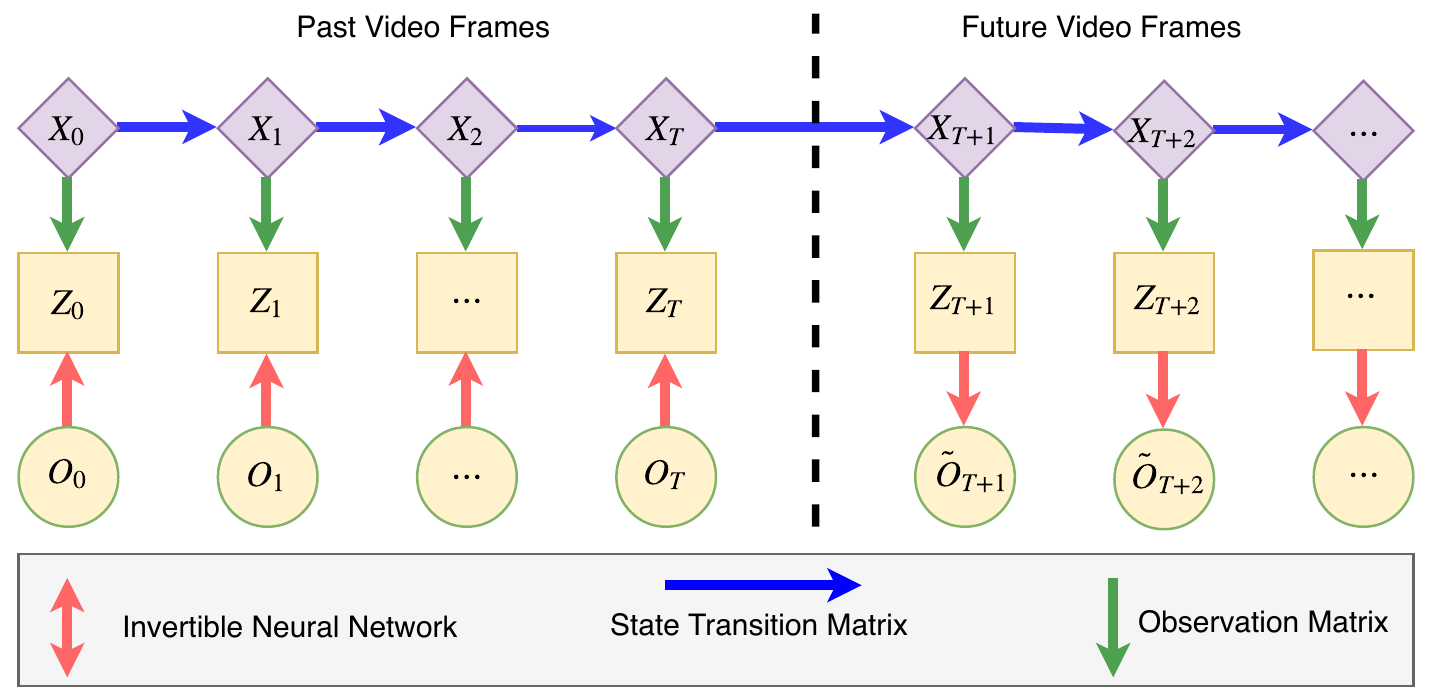}
\caption{We model video frame generation as a nonlinear dynamic system. The task is to find a suitable invertible linear embedding which encodes pixel-space observations $o_t$ as $z_t$. We treat $z_t$ as observations of a Markov process and solve for the initial latent state $x_0$. The invertible neural network, state transition matrix, and observation matrix are all learned parameters. This enables exact maximum likelihood learning without reconstruction, adversarial, or lower-bound losses in the image domain.}
\end{figure}

The grand vision of representation learning is to understand how useful encodings can be learned. Although there is no consensus as how this should be done, the video prediction objective offers a well-posed unsupervised task that may provide insight into how these productive representations may be learned and may produce them itself \cite{mathieu2015deep}.

 Video prediction can be modeled as a probability distribution over future frames. This stochasticity presents a challenge to prediction tasks. Consider a video of a falling leaf - it may float in one direction or another with little indication of future direction. To approximate the true distribution of future frames, some approaches minimize pixel-wise mean squared error (MSE) reconstruction loss. Others optimize a lower bound, typical of variational auto-encoders (VAEs). Still others use generative adversarial (GAN) discriminators to approximate the likelihood in the data domain. However, none of these approaches model the true distribution.

Our main contribution, an invertible linear embedding (ILE), combines invertible neural networks and a latent linear dynamical system to explicitly model the true distribution. By leveraging an invertible function approximator and the change of variables formula, frame prediction likelihood can be precisely equated with the likelihood of an observation from a linear dynamical system.

\subsection{Related Work}
Modern approaches to frame prediction use advances in neural network research to make state-of-the-art gains in predictive performance, our workincluded. 
\cite{oh2015action} and \cite{chiappa2017recurrent} use an action-conditioned deep auto-encoder to estimate the frames of Atari games. \cite{reda2018sdc} uses a network to predict optical flow that is then used to warp frames into the future.
\cite{DBLP:journals/corr/abs-1804-01523} attempts to model the true distribution by fusing VAEs with GANs, using the VAE approach to discourage mode collapse in GANs, and the GAN discriminator to overcome the lower-bound approximation in VAEs. 
Similar to our work, \cite{watter2015embed} and its extension \cite{banijamali2017robust} use locally-linear dynamical models that use a latent structure, but ultimately rely on a variational lower bound to approximate the posterior. \cite{mathieu2015deep} and its extension \cite{lotter2016deep} use image gradient losses and adversarial training to achieve stable and crisp results. Our experiments compare against this particular algorithm. These works differ in both their network topology and the loss functions they employ to approximate the true distribution, but all use some form of reconstruction error as a regularization. We explore the impact this common assumption has on maximum likelihood estimation later in this work.

Conceptually, modeling a nonlinear dynamic system as an tuple of an encoding function and a linear (possibly time-invariant) dynamic system is known in control literature as a Hammerstein-Wiener block model\footnote{Technically the model presented here is a Wiener model, but we consider the connection to the generalized model in the literature to be important.} \cite{janczak2004identification}. This literature has historically focused on low-dimensional systems using methods which do not scale well to high dimensional systems like video. We extend it here with neural network techniques for high dimensional systems such as image sequences.

\section{Method}
Formally, we consider a video sequence as an ordered tuple of $T$ frames, each denoted as $o_t$. The abstract problem of video extrapolation is to learn the conditional distribution over future frames, given past frames:
\begin{align*}
    p(o_{t}\ |\ o_{t-1},\ \dots,\ o_{0})
\end{align*}
This distribution is extremely complicated with no closed form to tractably sample, score, or approximate directly. In lieu of direct approximation, we consider transformed frames $g_\theta(o_t) = z_t$, where $g$ is a neural network parameterized by $\theta$, which we refer to as \emph{embeddings} or \emph{encodings}:
\begin{align*}
     p(g_\theta(o_{t})\ |\ o_{t-1},\dots,\ o_0)
     = p_\theta(z_{t}\ |\ z_{t-1},\dots,\ z_0)
\end{align*}
Provided that $g_\theta$ is sufficiently expressive and invertible, we can define an equivalence between a typically tractable distribution over observations $p_\theta$ parameterized by $\theta$ and the true distribution over frames $p$ using a change of variables:
\begin{align*}
    p(o_{t}\ |\ o_{t-1},\ \dots,\ o_0)
    = p_\theta(z_{t}\ |\ o_{t-1}, \dots,\ o_0)\ \lvert\ \det \frac{\partial z_t}{\partial o_t}\ \rvert
\end{align*}
This equality allows us to learn the true distribution using a maximum likelihood objective:
\begin{align}
\label{eqn:truedistro}
\max_{\theta}\ p_{\theta}(z_{t}\ |\ o_{t-1},\dots,\ o_0)\ \lvert\ \det \frac{\partial z_t}{\partial o_t}\ \rvert
\end{align}
In this work, we use an invertible neural network as the model class for $g_\theta(o_t) = z_t$ and a linear time-invariant dynamic system (LTI) to define the tractable likelihood $p_\theta$. To our knowledge, this is the first work to demonstrate successful learning in reversible flow networks using an LTI prior and one of the few works in video frame extrapolation to avoid making any assumptions about the data distribution.
%
\begin{spacing}{2}
\begin{algorithm}[tb]
\linespread{1.55}
\caption{Invertible Linear Embedding}
\label{alg:algorithm}
\begin{algorithmic}[1] 
\STATE \textbf{Returns the following:}
\STATE  \ \ \ \ $g_\theta$: a learned invertible neural network
\STATE  \ \ \ \ $A$: a learned state transition matrix
\STATE  \ \ \ \ $C$: a learned observation matrix
\WHILE{$\mathcal{L}$ is not minimized}
\STATE Sample $o_0, \ldots, o_{T-1}$ frames
\FOR{$t = 0, \cdots, T-1$ }
   \STATE $z_t = g_\theta^{-1}(o_t) \in \mathbb{R}^{D}$
   \STATE $s_t = \lvert \det \frac{\partial z_t}{\partial o_t} \rvert$
\ENDFOR
\\
\STATE $\mathcal{Z} = \begin{bmatrix} z_0 \\ z_1 \\ z_2 \\ \vdots \\ z_{T-1} \end{bmatrix}$ \hspace{.1cm} $\mathcal{O} = \begin{bmatrix} C \\ CA \\ CA^2 \\ \vdots \\ CA^{T-1} \end{bmatrix}$\\
\STATE $x^*_0 = \mathcal{O}^+\mathcal{Z}$\\
\STATE $\hat{\mathcal{Z}} = \mathcal{O}x_0^*$\\
\STATE $\gamma = \frac{1}{D} \lVert \mathcal{Z} \rVert_1$\\
\STATE $\mathcal{L} = \frac{1}{2}\lVert \gamma^{-1} (\mathcal{Z} - \hat{\mathcal{Z})} \rVert_2^2 + \sum_{t}^T[ \log s_t ] - \log \gamma$\\
\STATE Take gradient step in $A$, $C$, $\theta$ to minimize $\mathcal{L}$\\
\ENDWHILE
\STATE $o_{T} = CA^Tx_0^*$
\end{algorithmic}
\end{algorithm}
\end{spacing}

\subsection{Reconstruction Error and Implicit Assumptions}
To distinguish between a large class of prior work and our contribution, we highlight a distinction between a common candidate objective function and the true distribution described in Equation \ref{eqn:truedistro}. A common framework for video prediction involves learning an encoding function $e_\theta$, a separate decoding function $d_\theta$, and a transition function $f_\theta$. Learning the decoder has the practical purpose that the system avoids perfectly predictable but not particularly useful minima such as $e_\theta = 0$. A typical loss is usually defined:
\begin{align*}
\min_\theta\ \alpha \lVert f_\theta(e_\theta(o_t)) - e_\theta(o_{t+1}) \rVert + \beta \lVert d_\theta(e_\theta(o_t)) - o_{t}\rVert
\end{align*}
When using $L^2$ as the norm, minimizing this candidate objective function is equivalent to maximum log likelihood learning under three assumptions. First, that the conditional distribution of the error \textit{of the embedding} is an isotropic Gaussian.
$$p_\theta(e_\theta(o_{t+1}) | e_\theta(o_t)) = \mathcal{N}(e_\theta(o_{t+1}), \alpha^{-1})$$
Second, that the input images are isotropic Gaussian with a mean defined by the decoder 
$$p(o_t | e_\theta(o_t)) = \mathcal{N}(o_t, \beta^{-1})$$
Third, that the determinant of the encoder's Jacobian is 1.
If our observations are image pixel intensities, these may not hold. While the first assumption is valid (given a sufficiently expressive encoder), the second and third are not. The term $\lVert d_\theta(e_\theta(o_t)) - o_{t}\rVert$, which we call \emph{reconstruction loss}, compares an observation $o_t$ with its reconstruction $d_\theta(e_\theta(o_t))$ using pixel-wise mean-squared error,  known to perform poorly in the case of translations and brightness variations.  More generally, it completely ignores valuable higher-order information in images: pixel intensities are neither independent nor do they share the same variance. The third assumption is likewise almost certainly not true for traditional auto-encoders. Put simply, this loss implies false assumptions and results in a different objective than the one we would truly like to minimize. 

In GAN approaches, the error between the true and approximate distribution is theoretically bounded by the complexity of the discriminator which acts as a data-driven direct approximation to $p(o_t | \cdot)$. In practice, adversarial losses are difficult to train and are generally used as an additional regularizing term in loss functions which make similar simplifying assumptions about the data distribution \cite{mathieu2015deep}.

\subsection{Background: Invertible Networks}
Recent work on invertible networks, also known as reversible flows, are a relatively new approach to deep generative modeling~\cite{dinh2014nice,dinh2016density,kingma2018glow} which introduces techniques for learning exactly invertible neural networks.  This work represents the backbone of our method. We consider a generative model using a known parameterized distribution $p_\theta(z)$ and a deterministic function $g_\theta(z)$:
\begin{align*}
    z \sim p_\theta(z), & & o = g_\theta(z),
\end{align*}
w\textbf{}here $g_\theta(z)$ has the compositional form 
$$g_\theta^{(N)}(g_\theta^{(N-1)}(\cdots g_\theta^{(0)}(z)\cdots))$$ in which each successive layer operates on the output of the layer before (abbreviated notationally as $g_\theta^i(h_{i-1})$). The change of variables formula enables us to relate:
\begin{align*}
    \log p(o) = \log p_\theta(g^{-1}(z)) + \sum_{i=0}^{N}\ \lvert\  \det \frac{\partial h_i}{\partial h_{i-1}}\ \rvert,\textbf{}
\end{align*}
with $h_0=o$ and $h_N=\textbf{}z$. Because $p_\theta$ is tractable, we need only for the determinant of each layer's Jacobian to be tractable to efficiently compute the density $\log p(o)$.

Borrowing on the early work in this field, we consider the following technique for $g_\theta^{(i+1)}(h_{i})$ called an \textit{affine coupling} which makes this determinant easy to compute:
\begin{align*}
    h_{i}^\mathrm{left}, h_{i}^\mathrm{right} &= \mathrm{split}(h_{i})\\
    s_i &= f_i( h_i^\mathrm{left}) \\
    h_{i+1}^\mathrm{right} &= s_i \odot h_{i}^\mathrm{right} + b_i(h_i^\mathrm{left})\\
    h_{i+1} &= P_i \begin{bmatrix} h_{i}^\mathrm{left} \\
    h_{i+1}^\mathrm{right} \end{bmatrix}
\end{align*}
where $h_i \in \mathcal{R}^D$ is the layer input, $\odot$ is the element-wise product, $f_i$ and $b_i$ are arbitrary neural networks (not necessarily invertible), and $P_i$ is a unimodular matrix which mixes elements between the two halves of $h_i$. Although this may seem intimidating, the computation is straightforward: using half of the layer's input we learn to produce an affine transformation to apply to the other half. This operation is invertible, and the log-determinant of this layer is simply:
\begin{align*}
    \log\ \lvert \det \frac{\partial h_i}{\partial h_{i-1}} \rvert = \log\ \sum_{j=0}^{D/2} \lvert s_{ij} \rvert.
\end{align*}
The log-determinant of the entire encoding function is the sum of these terms for all layers $i$:
\begin{align*}
    \log\ \lvert \det \frac{\partial g^{-1}_\theta(o_t)}{\partial o_{t}} \rvert = \log\ \sum_{i=0}^N \sum_{j=0}^{D/2}\ \lvert s_{ij} \rvert.
\end{align*}

Taken together, this functional form enables us to define our decoding function $g_\theta(z)$, its exact inverse, and an efficient computation for the log-determinant of its Jacobian. 
 
\subsection{Background: Linear Latent Prior}
Recall that in our primary objective function in addition to being able to learn $g_\theta(z)$, we must also be able to define a tractable computation for the distribution $p_\theta(z_t | z_{t-1}, \dots z_0)$. In this section we introduce linear time-invariant systems as this density function. A natural model for the evolution of a vector-valued observation is that of a linear dynamic system:
\begin{align*}
    x_{t} = Ax_{t-1} \qquad
    z_{t} = Cx_{t-1} + \gamma_{t-1} \qquad \gamma_{t-1} \sim \mathcal{N}(0, I)
\end{align*}
Where $x_t$ represents the hidden state, and $z_t$ the observation at that hidden state. In this work, we assume that this system is \emph{time-invariant}; however, we note that it is possible to extend this model to not only include time-varying dynamics, but also inputs, process noise over the hidden state, or noise distributions with different distributions, but omit them in this work for simplicity.

Linear Time Invariant (LTI) systems define a conditional distribution with tractable density over observations:
\begin{align*}
    p_\theta(z_t | z_{t-1}, \dots, z_{0}) 
    &= \mathcal{N}(CA^t{x^*}_{0}, I)
\end{align*}
where $x^*_0$ is the result of optimal latent state inference. For LTI systems, this is the result of a Kalman filter when conditioned on only past observations, and the Kalman smoother when conditioned on both past and future observations \cite{welch1995introduction}. Although many algorithms \cite{thrun2005probabilistic} exist to compute the optimal smoothing estimate $x^*$ they can all be shown to produce the same least squares estimate:
\begin{align*}
    x^*_0 &= \arg \max_{x_0} \left\lVert \begin{bmatrix} z_0\\ z_1 \\ \vdots\\ z_{T-1} \end{bmatrix} - \begin{bmatrix} C\\ CA \\ \vdots\\ CA^{T-1} \end{bmatrix}x_0 \right\rVert_2^2\\
    &= \arg \max_{x_0} \left\lVert \mathcal{Z} - \mathcal{O}x_0 \right\rVert_2^2\\
    &= \mathcal{O}^+\mathcal{Z}
\end{align*}
The tractable density function is:
\begin{align*}
    p_\theta(z_t | o_{t-1}, \dots, o_{0}) &= \mathcal{N}(CA^t\mathcal{O}^+\mathcal{Z}, I)
\end{align*}
where we use $M^+$ as the pseudoinverse of $M$. The efficiency and optimality of hidden state inference is one of the motivating factors behind our choice of a linear model for the latent evolution of embeddings. 

Using a linear dynamic system as the transition model for our system introduces two major assumptions. The weaker assumption is the Markov property, that future observations are independent of past observations when conditioned on the hidden state. The stronger assumption is that the future hidden states are a linear mapping from past states and that this mapping remains constant through time. Although the linear dynamics prior may seem quite restrictive, LTI systems are surprisingly expressive and have been shown to model the latent dynamics of many high dimensional models  \cite{brunton2016discovering,lusch2018deep}. A key theoretical insight in control literature proves the existence of an infinite dimensional linear operator, the Koopman operator, for some nonlinear projection of all nonlinear dynamic systems \cite{koopman1931hamiltonian}. When choosing a large latent hidden dimension, we are approximating this infinite-dimensional operator. So while the modeling assumption made by a linear dynamical prior is almost certainly not true, a large enough state space is a good approximation and will demonstrate the viability of ILE for difficult non-linear systems. Future work could explore options for more expressive yet tractable time-variant dynamic system models.

\subsection{Invertible Linear Embedding}
We now present our primary contribution: the invertible linear embedding.

Using an invertible neural network as our encoding and decoding function $g_\theta(o)$ and $g_\theta^{-1}(z)$, and an LTI dynamic system as described for the conditional distribution, we can derive our final loss function:
\begin{align*}
    \mathcal{L} &= -\log p_\theta(o_t\ |\ o_{t-1}, \dots, o_{0})\\
    &= -\log[\ p_\theta(g^{-1}_\theta(o_t)\ |\ o_{t-1}, \dots, o_{0})\lvert \det \frac{\partial g^{-1}_\theta(o_t)}{\partial o_t} \rvert]\\
    &= -\log\ p_\theta(z_t\ |\ o_{t-1}, \dots, o_{0}) - \log\ \lvert \det \frac{\partial z_t}{\partial o_t} \rvert\\
    &= -\log\ \mathcal{N}(CA^t\mathcal{O}^+\mathcal{Z}, I) - \log\ \lvert \det \frac{\partial z_t}{\partial o_t} \rvert
\end{align*}
Which results in:
\begin{align}
    \mathcal{L} &= \frac{1}{2}\lVert \mathcal{Z} - \mathcal{O}\mathcal{O}^+\mathcal{Z} \rVert_2^2 - \log\ \sum_{i=0}^N \sum_{j=0}^{\frac{D}{2}}\ \lvert s_{ij} \rvert
\end{align}
When minimized using sufficiently expressive $g_\theta(z), A$, and $C$ parameters, this loss function corresponds to \text{exact} maximum likelihood model of a video sequence which is assumed to have latent linear dynamics.

We can describe the function of these two terms intuitively. The first term (the \textit{predictive error}) is the result of encoding each frame independently, solving for the best possible LTI dynamic system trajectory, and applying gradient descent to minimize any error. The more the embeddings behave as a linear system, the lower the predictive error. The second term (the \textit{log-determinant}) encourages the embeddings to be large, preventing the first term from collapsing to easy-to-predict but useless trajectories such as $z_t = 0$. Although it may seem like a strange regularization to ``maximize the embedding values", the application of change of variables and strict invertibility ensures that this is \text{the} correct way to learn a mapping between our assumed latent model, and the true observations in image space.

\subsection{Stability and Parameterization of \texorpdfstring{$A$}{A}}
Given the numerical instability induced from computing a least-squares solution in our training loop, the parameterization of the linear dynamic system is of critical concern. In particular, we must parameterize the learning method to maintain stable state transition matrices $A$. A stable discrete-time linear dynamic system is one where $\sigma(A) < 1$, so $A^t$ does not explode for large $t$. 

One feature of LTI systems that we can exploit to ensure training stability is that, for a given state-space parameterization $A, C$, there exist an infinite number of equivalent parameterizations that correspond to the same system and thus produce the same observation sequences. This can be explained intuitively: one can rotate the hidden state space by some transformation $T$, evolve the state in this transformed space before de-transforming observations with $T^{-1}$. In practice, this means we can consider any parameterization for $A$ which has the eigenvalues of the true system if we learn a dense $C$ matrix.

Because the stability of a linear dynamic system is characterized by the magnitude of the eigenvalues of $A$, 
we can choose $T$ so it is easy to compute and restrict these values.
If the true system $A^* = Q\Lambda Q^{-1}$ with a complex diagonal matrix of eigenvalues $\Lambda$, then we choose $T = Q$ implying that we learn $A = \Lambda$. This both decreases the number of learnable parameters in $A$ while also making enforcing stability relatively trivial.

\subsubsection{Jordan Normal Form}
The primary issue with learning $A = \Lambda$ is that $\Lambda$ as the eigenvalues of a real matrix will come in complex conjugate pairs\footnote{If we assumed that $A^*$ was symmetric, the eigenvalues would have no imaginary components and we could instead simply learn a diagonal real matrix $\Lambda$.}. However, Real Jordan Normal Form (JNF) \cite{} offers a simple solution. By splitting the real and imaginary parts, we can construct an all-real matrix for which matrix multiplication simulates complex multiplication with this constraint. The following represents a $4\times4$ example:
    \begin{align*} A = \begin{bmatrix}
    \alpha_0 & \beta_0 & 0 & 0\\
     -\beta_0 & \alpha_0 & 0 & 0\\
     0 & 0 & \alpha_1 & \beta_1\\
     0 & 0 & -\beta_1 & \alpha_1
     \end{bmatrix} \end{align*}
    
This form does have its drawbacks. In scenarios where any imaginary components are actually zero, then there should be an additional unique real component. Although somewhat inelegant, the negative impact of this scenario can be mitigated by simply increasing the dimensionality of $A$. Additionally if $\alpha_0 = \alpha_1$ and $\beta_0 = \beta_1$ true Jordan blocks should additionally have a one in the off-diagonal corresponding to eigenvalues with multiplicity greater than one. In practice this is not an issue as it is difficult to produce \textit{exactly} identical eigenvalues.

Although there are many ways to ensure that the magnitude of each eigenvalue in the JNF does not exceed $1$, we found the following reparameterization to be effective, using $\theta_\alpha$ and $\theta_\beta$ as vectors of unconstrained real-valued parameters to produce the vectors of constrained real and imaginary components $\alpha$ and $\beta$:
\begin{align*}
\alpha &= max((1 - \epsilon) - \lvert \theta_\alpha \rvert, 0)\\
\beta &= max(1 - \lvert \theta_\beta \rvert, 0) * \sqrt{1 - \alpha^2}
\end{align*}
where $\epsilon = 10^{-14}$. This particular transformation ensures that every unique real parameter pair $\theta_\alpha, \theta_\beta$ corresponds to a unique complex eigenvalue. The small epsilon subtraction ensures that we never compute $\sqrt{0}$. In our implementation, $\epsilon$ is chosen such that when we compute $\alpha$ and $\beta$ with double precision, and then cast to single precision floating point we avoid $\sqrt{0}$ and allows $\alpha = 1$. 

\subsection{Addressing the Scale Ambiguity}
When learning both the encoding function and the dynamic system parameters simultaneously, there is an ambiguity between the scale of the embedding and the scale of the dynamic system when the covariance is learned. As an illustrative example, consider the following system:
    \begin{align*}
    y_t = \gamma_t f_\theta(o_t) \qquad \hat{y}_t = \gamma_t C x_t
    \end{align*}
A scaling ambiguity occurs when we try to learn the covariance of the error in addition to the other parameters of our network, i.e when the predictive loss becomes:
\begin{align*}
\log p(y_t | y_{t-1}) &\propto (y_t - \hat{y}_t)^T\Sigma(y_t - \hat{y}_t) \\
&= (\gamma_t f_\theta(o_t) - \gamma_t C x_t)^T\Sigma(\gamma_t f_\theta(o_t) - \gamma C x_t)\\
&= \gamma_t^2 (f_\theta(o_t) - C x_t)^T\Sigma(f_\theta(o_t) - C x_t)
\end{align*}

The $\gamma_t^2$ term, which induces downward pressure on the embedding magnitudes when $\Sigma$ is constant, can be absorbed as a learned $\Sigma$ adjusts during training. This effectively removes its impact, but leaves behind the upward pressure on magnitudes from the $\log \gamma_t$ term, which will result in the system maximizing $\gamma_t$, rather than prediction error. In practice, this results in a runaway scale of the embeddings and numerical issues.

To address this we model the $\gamma^{-1}$ as another layer in our invertible network which we simply adds another term to our loss function:
\begin{align*}
\mathcal{L} = \log p(\gamma_t^{-1}y_t | \cdot ) + \log \lvert \det \frac{\partial y_t}{\partial o_t} \rvert - \log \gamma_t
\end{align*}

In practice, we found that this adjustment improved training stability even when the covariance is held constant during training. Although $\gamma_t$ could be learned, we used $\gamma_t =  \frac{1}{N}\lVert y_t \rVert_{1}$. We also found that the $L^1$ norm performs better than the $L^2$ norm\footnote{Presumably because it better propagates small gradients in each dimension of $y_t$.}.

\section{Experiments and Results}

\begin{figure*}[ht]
\centering
\includegraphics[width=\linewidth]{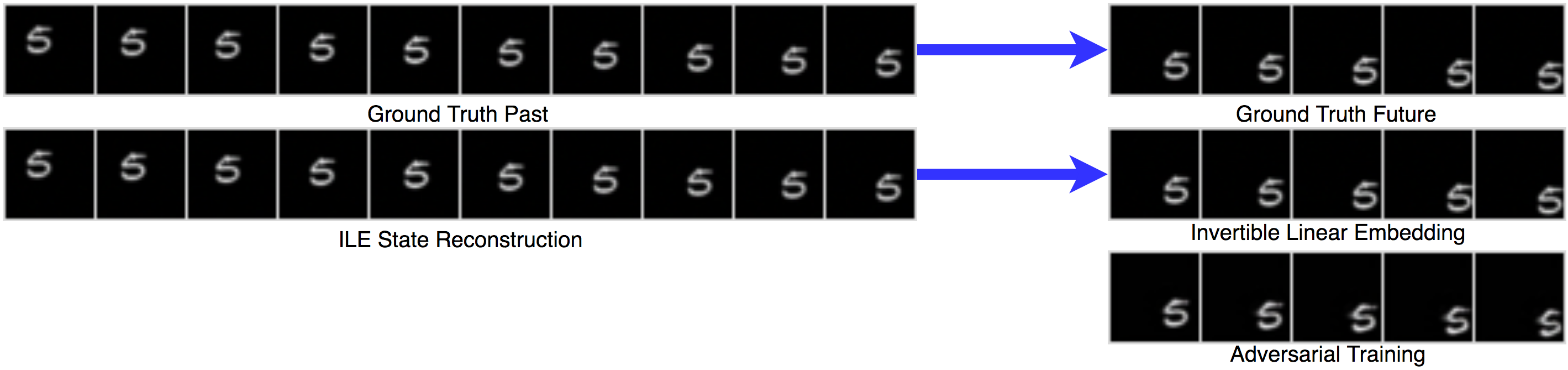}
\caption{The Bouncing-MNIST dataset, modeling elastic collisions which preserve object shape.}
\end{figure*}

\begin{figure*}[ht]
\centering
\includegraphics[width=\linewidth]{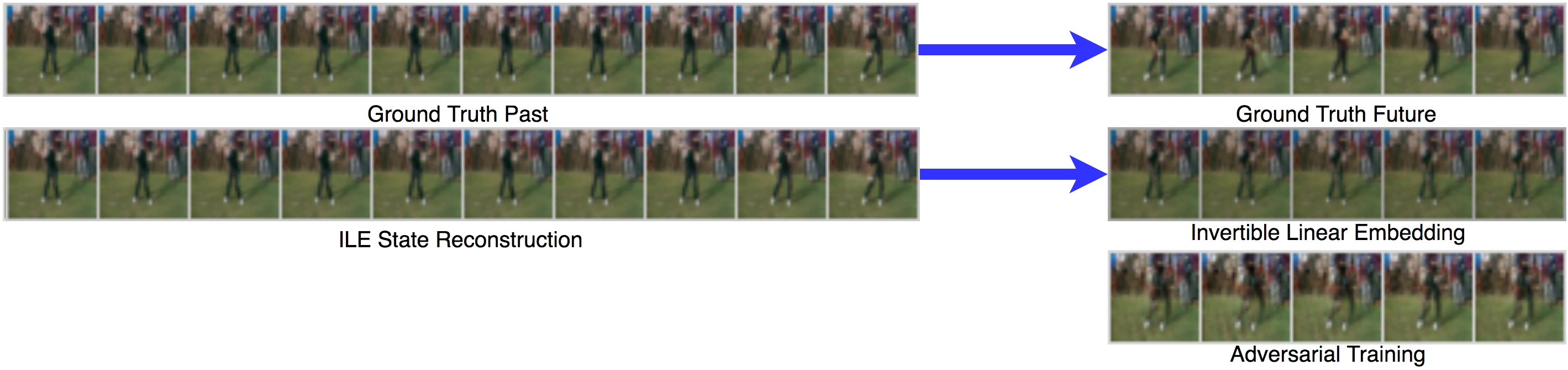}
\caption{The UCF Sports Action dataset, modeling the progression of a golf swing.}
\end{figure*}

\subsection{Datasets}
We show results for both a synthetic and realistic dataset. The synthetic data, entitled Bouncing-MNIST, is generated using the Moving Symbols algorithm, a published benchmark designed to support the objective study of video prediction networks \cite{wang2004image,szeto2018dataset}. Each video sequence samples an MNIST digit, assigns it an initial trajectory, and simulates elastic collisions with the image boundary.

The realistic sequences are sampled from UCF Sports Action \cite{rodriguez2008action,soomro2014action}. This dataset contains video sequences of various sports such as diving, running, horseback riding, and golfing.

\subsection{Network Topology}
Our network is most similar to that used by \cite{kingma2018glow}, but without $1\times1$ convolutions, or the act-norm operation. We used $4$ blocks of $10$ affine-coupling layers each, where each block has an early connection out to the final embedding. Our non-invertible networks used at each step of flow were simple 3-layer networks of $3\times3$ convolution with two output channels for the affine transformation parameters and $512$ channels in the center.
For comparison, we implement the adversarial training algorithm of \cite{mathieu2015deep}, which is known for its sharp image quality.

\subsection{Results}

\begin{figure}[H]
\centering
 \resizebox{\linewidth}{!}{%
 \begin{tabular}{| c | c c | c c |} 
 \hline
 \multirow{2}{4em}{Method} & \multicolumn{2}{|c|}{First Frame} & \multicolumn{2}{|c|}{Fifth Frame} \\
 & PSNR & SSIM & PSNR & SSIM \\ 
 \hline
 Invertible Linear Embedding & \textbf{23.5} & 0.92 & \textbf{17.4} & 0.69 \\ 
 Adversarial Training & 20.6 & \textbf{0.95} & 12.1 & \textbf{0.83} \\
 Last Input & 17.3 & 0.76 & 14.5 & 0.67 \\
 \hline
\end{tabular}
}
\caption{Peak Signal-to-Noise Ratio (PSNR) and Structural Similarity (SSIM) scores, taking the mean over 100 held-out test sequences. We generate future frames $o_1, o_2, ..., o_5$ and calculate scores on $o_1$ and $o_5$ to measure both immediate and longer-horizon prediction quality. We again note that our approach does not explicitly minimize the mean squared error between predicted frames and ground truth.}

\end{figure}

We evaluate our algorithm by comparing against adversarial training in three ways: qualitatively through examples, with peak signal-to-noise ratio (PSNR), and with the structural similarity (SSIM) index \cite{wang2004image}. Statistical results are reported on the synthetic dataset. 

Although adversarial training has a slight advantage in SSIM, the ILE algorithm outperforms it in PSNR. The difference is especially pronounced over a longer time horizon. Adversarial training maintains crisp shapes, yet lacks accurate motion projections over even moderate time horizons. After five frames it performs significantly worse than the naive baseline. ILE maintains a reasonable representation of the digit shape, and excels at motion projection over a long time horizon, even accurately predicting bounces off image boundaries. This suggests that the nonlinear dynamic system is being fit quite well.\footnote{The adversarial training PSNR scores are lower than those reported in \cite{mathieu2015deep} because the synthetic dataset has much more motion than the UCF-101 dataset, which the original paper used as a benchmark. In our tests, digit velocity is up to 3 pixels/frame in each direction. However, the high velocity is intentional; a quality benchmark for video prediction should use sequences where motion is noticeable.}

While adversarial training performs well on sequences where the motion is strictly linear, such as those pictured, it performs poorly in motion that is nonlinear in pixel space. For example when the digit bounces off a wall or when a golf club accelerates in the frame. In contrast, ILE models all motion sequences well, suggesting better generalization ability.

\section{Directions for Future Work}
Our work moves toward exact maximum likelihood optimization to improve performance in video prediction. We present here what we consider to be natural next steps, and the implications they might have.

\textbf{Action-conditional and time-variant models.} By extending the model of the hidden dynamical system to include an action $u$ and linear mapping $B$ it becomes possible to use ILE for model based reinforcement learning and optimal control. Additionally a simple extension to our model which may prove promising is to learn a time or state-conditional state-transition matrix $A_t$ in lieu of the constant $A$ presented. This particular extension could be done using any standard autoencoding neural network architecture as a JNF state transition matrix is diagonal and invertibility is not a requirement. Although time-varying linear dynamic systems are more difficult to analyze, they are models of much greater capacity and therefore could be better suited for difficult problems that would require infinite, or near-infinite dimensional state dimensions.

\textbf{Scaling to larger frame dimensions.}
Invertible networks, perhaps as a direct result of the difficult task of modeling the entire unknown distribution over video frames, are large and difficult to train. In particular, memory usage in our model even for these relatively small frame sequences was a computational constraint. Architectural improvements such as those recently proposed by \citeauthor{grathwohl2018ffjord} could extend our results into images approaching modern video resolutions.

\section{Conclusion}
We have presented the invertible linear embedding, which provides exact maximum likelihood learning of video sequences.  Our key contribution is to combine invertible networks with linear dynamical systems. While images sequences may lie on a complex probability manifold in high-dimensional space, an invertible network coupled with a change of variables learns how to properly map that manifold of probability to the well-behaved conditional Gaussian created by a linear dynamic system. By formulating this with a single learning objective, we arrive at an elegant joint optimization problem.  The primary advantage of this approach is that we avoid making any assumptions about the distribution of the input domain.

In future work we believe even better qualitative performance can be had as more becomes known about optimization and training of invertible networks.

\bibliographystyle{named}
\bibliography{ijcai19}

\end{document}